# HazardFuse: Hazard-Index-Guided Band Attention and Spatial Embedding Fusion Network for Leached-Metal-Associated Hazard Mapping[1]

Salman Karim, Bahareh Nojabaei

***Abstract*—Coal combustion residual (CCR) landfills and impoundments can leach potentially toxic metals into groundwater, but monitoring wells observe the resulting hazard only at discrete points. This study presents HazardFuse, a hazard-index-guided band attention and spatial embedding fusion network that extends well observations into continuous maps of leached-metal-associated hazard. A composite hazard index (HI) of the constituents measured in 19 monitoring wells was kriged and divided into lower- and higher-HI classes, which served as supervision. The predictors formed a 63-channel stack of Sentinel-2 reflectance, vegetation and biophysical indices, spatio-temporal variation metrics, and soil, climatic, hydrologic, topographic and land-cover variables. HazardFuse weights the bands of the target pixel through band attention, encodes the 21 × 21-pixel neighborhood through a spatial-embedding branch and combines the two through a learned gate. At the former Glen Lyn Power Station, Virginia, USA, it was compared with ten machine-learning and deep-learning benchmarks under one spatial block split with buffer and one calibration rule. Over three training runs, HazardFuse reached a mean test accuracy of 0.973 ± 0.002 and Cohen's κ of 0.929 ± 0.005, with precision, recall and F1 of 0.951, 0.943 and 0.947 for the lower-HI class and 0.980, 0.983 and 0.982 for the higher-HI class. It ranked first in every run and made 22–39% fewer errors than the best benchmark of each run. Ablation showed that neighborhood context and band attention contributed most, and that soil, Sentinel-2, vegetation indices and rate-of-change inputs were the most informative feature groups. The maps indicate where additional groundwater sampling around CCR units would be most useful.**

***Index Terms*—coal combustion residuals (CCR), deep learning, pixel-patch gated fusion, groundwater hazard index, leached metals, Sentinel-2, band attention, spatial embedding.**

## I. Introduction

Coal combustion residuals (CCR), including fly ash and bottom ash from coal-fired power generation, contain inorganic constituents such as arsenic, selenium, boron, molybdenum and lithium which can be mobilized through water–ash interaction and transported into groundwater and adjacent surface water [1], [2]. Legacy CCR landfills and surface impoundments are therefore monitored with networks of groundwater wells, which support closure and corrective-action decisions under the federal CCR Rule [3], [4] code and state programs such as the Virginia Waste Management Act [5], [6]. Monitoring wells provide direct subsurface observations, but they are spatially discrete. On their own, they cannot describe the continuous pattern of hazard between sampling locations.

Earth-observation and ancillary environmental data offer a complementary view of the landscape around CCR units. Sentinel-2 multispectral imagery, which includes three red-edge bands [7], captures spatial and temporal variation in vegetation and surface properties that has been associated with metal-induced vegetation stress [8], [9], [10]. Terrain, soil, climatic, hydrologic and land-cover variables describe environmental controls on both vegetation condition and contaminant transport [11]. These data do not measure dissolved metals, and similar spectral responses can arise from drought, nutrient limitation, disease or land management [9]. They instead provide a heterogeneous predictor space from which spatial associations with groundwater-derived hazard can be learned.

This formulation is established in groundwater-quality mapping at regional to global scales; for example, machine learning on climatic, soil, geological and topographic predictors has been used to map the probability of elevated arsenic in groundwater worldwide [12]. Such studies generally apply pixel-wise models to point observations at coarse resolution. The setting of a CCR facility differs in three respects: the monitoring network is small and spatially clustered; the target reflects several co-occurring constituents rather than one; and the relevant surface signal is expressed over tens to hundreds of meters, so that neighborhood context carries information.

Learning from a heterogeneous multisource stack at site scale raises four requirements that existing model families do not meet.

*1) Explicit relevance of heterogeneous input bands:* The input bands represent different physical domains (reflectance, vegetation indices, biophysical parameters, temporal metrics, soil, climate, hydrology, terrain and land cover). They differ in units, dynamic range and spatial smoothness and are unlikely to contribute equally. Channel-attention mechanisms such as the squeeze-and-excitation (SE) block [13] and CBAM [14] weight learned convolutional feature maps, and hyperspectral networks apply attention to contiguous spectral bands [15], [16]. Neither assigns relevance directly to the original, physically interpretable bands at the target location.

*2) Neighborhood spatial context:* Environmental patterns are spatially autocorrelated, so the state of a location depends partly on its surroundings [17]. Random forest [18] is robust to redundant and mixed-type predictors and is widely used in remote sensing [19], but it classifies per-sample feature vectors and learns no neighborhood representation. Patch-based convolutional neural networks (CNNs) classify the center pixel of a local image patch and address this limitation [20].

[1] This work was supported by the U.S. Department of Energy, National Energy Technology Laboratory, under Award DE-FE0032184. (Corresponding author: Salman Karim.) Salman Karim and Bahareh Nojabaei are with the Department of Mining and Minerals Engineering, Virginia Tech, Blacksburg, VA 24061 USA (e-mail: salmankarim@vt.edu; baharehn@vt.edu).

*3) Adaptive combination of the two information streams:* Most multi-branch networks fuse their streams by concatenation or addition, with weights that are fixed after training [21], [22]. Gating mechanisms instead compute sigmoid-activated, element-wise weights from the input [23], [24], [25]; the same principle underlies channel recalibration [13] and gated fusion of multi-level features in segmentation [26]. In remote sensing, the fusion strategy has a large effect on multimodal classification accuracy [27].

*4) Compatibility with sparse, point-derived supervision:* Hazard labels are derived from a limited number of wells rather than from dense annotated masks. Encoder–decoder segmentation networks such as U-Net [28], U-Net++ [29], ResU-Net [30], [31], SE-residual U-Nets [32], DeepLabv3+ [33] and SegFormer [34] have hundreds of thousands to millions of parameters and are designed for fully labeled images. A compact center-pixel classifier matches the structure of the labels more directly.

A further challenge is to define a target that represents the combined occurrence of several constituents. Building on the Hazard Quotient–Hazard Index (HQ–HI) method of the preceding study [11], this work uses a composite groundwater HI as the reference. Constituent-specific HQs standardize concentrations relative to reference thresholds, and the HI sums them, following the additive formulation used in risk assessment [35]. The HI used here is a study-specific classification construct and not a formal human-health or ecological risk index.

This study proposes HazardFuse, a hazard-index-guided band attention and spatial embedding fusion network for mapping leached-metal-associated hazard. Its three components correspond to requirements 1)–3): a band-attention branch that weights the raw center-pixel band vector; a spatial-embedding branch of convolutional and depthwise-separable convolutional layers [36] that encodes the surrounding patch; and a gated fusion module. Requirement 4) is met by the compact center-pixel formulation. HazardFuse is evaluated at the former Glen Lyn Power Station, Virginia, against ten benchmarks under a common spatial block design with buffer, which limits the optimistic bias that spatial autocorrelation introduces into random validation [17], [37].

The contributions of this study are as follows.

1. A composite, multi-constituent HI is used as spatial supervision for classifying leached-metal-associated hazard from a 63-channel multisource stack.
2. HazardFuse jointly learns the relevance of the target-location bands and the neighborhood context and combines them through a learned gate; its training and mapping procedure is given in Algorithm 1.
3. HazardFuse is compared with two machine-learning and eight deep-learning benchmarks under one spatial split, one rank-based calibration rule and a paired block bootstrap, using accuracy, Cohen's κ and class-wise precision, recall and F1.
4. Architectural and feature-group ablations and the learned band-attention weights quantify which components and inputs carry the predictive information.

## II. Related Work

### A. Machine Learning for Groundwater-Quality Mapping

Machine-learning models are widely used to predict groundwater-quality parameters from environmental covariates [12]. Tree-based ensembles, particularly random forest [18], dominate this literature because they handle mixed-type and correlated predictors [11], [19]. These studies relate point observations to covariates at the sample location; they treat each location independently of its neighborhood, and most model a single constituent.

### B. Remote Sensing of Metal-Induced Vegetation Stress

Metal uptake alters chlorophyll content, leaf structure and water status, and these changes are expressed most strongly in the red-edge region [38]. Red-edge and vegetation-index changes have been related to soil contamination in floodplains [8], and Sentinel-2 red-edge indices and their spatio-temporal variation have distinguished heavy-metal-stressed from unstressed crops [9]. Sentinel-2 vegetation and biophysical indices have also been related to leached-metal concentrations near coal ash impoundments [10]. These approaches are generally index-based and do not integrate the non-spectral controls that confound the vegetation signal.

### C. Deep Learning for Multisource Geospatial Classification

CNNs trained on patches centered on the target pixel outperform pixel-based networks on medium-resolution imagery [20], and patch-based CNNs applied to stacks of conditioning factors have outperformed conventional classifiers for landslide susceptibility [39]. For pixel-level mapping, encoder–decoder and atrous-convolution networks [28], [29], [33], [40] and Transformer-based segmenters [34] are standard. Two-branch networks extract features from different sensors before fusion [41], and the fusion strategy affects accuracy [27]. In all of these, the relevance of individual input variables is learned implicitly within shared convolutional filters.

### D. Channel Attention and Gated Fusion

The SE block [13] recalibrates channels with weights computed from globally pooled features, and CBAM [14] adds spatial attention. Spectral–spatial hyperspectral networks combine such attention with spatial encoders [15], [16], and SE blocks have been inserted into U-Nets [32]. These designs assume that channels are learned features or ordered spectral bands. Gated linear units [24] and gated multimodal units [25] learn how much each pathway contributes to a fused representation. The literature thus provides each element separately but does not combine them in one model for site-scale, multi-constituent hazard classification from heterogeneous inputs under sparse supervision.

## III. Study Area

### A. Site Description and CCR Management Units

The study area is the former Glen Lyn Power Station (37°22′12″ N, 80°51′48″ W), a coal-fired generating facility on the south bank of the New River in Giles County, southwestern

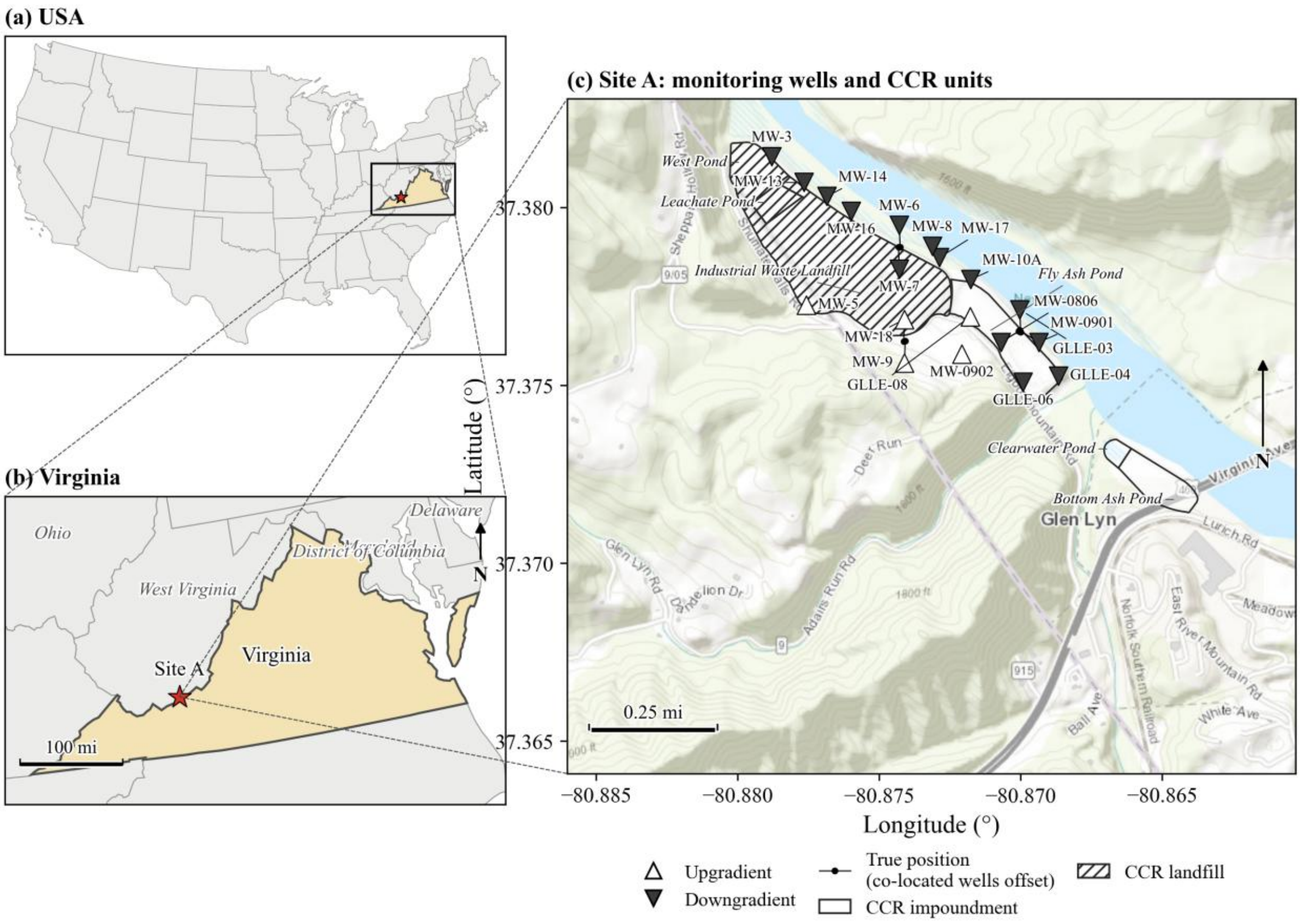

Fig. 1 : Study area. (a) Location in the conterminous United States and (b) in Virginia. (c) Site A, the former Glen Lyn Power Station (Site A), with the CCR landfill (hatched; Industrial Waste Landfill) and impoundments (West Pond, Leachate Pond, Fly Ash Pond, Clearwater Pond and Bottom Ash Pond) and the 19 groundwater monitoring wells (△ upgradient, ▼ downgradient). Co-located wells are offset from their true positions (dots)

Virginia, USA (Site A). The plant began operating in 1919 and reached a final capacity of approximately 338 MW before its retirement in May 2015. Over nearly a century of operation, CCRs were placed in several impoundments and a landfill within the New River floodplain (Fig. 1). The CCR units under the state solid waste permit SWP222 are the West Pond, a landfill, the Auxiliary Pond, a leachate pond and the Bottom Ash Pond [5], [42], [43]. The mapped Site A is a 212 × 288 pixel grid (pixel size 10 m, about 6.105 km²) containing the CCR units and the river corridors.

## *B. Groundwater Monitoring and Constituents*

Groundwater quality around the CCR units was characterized with data from 19 monitoring wells sampled between 2010 and 2014 [44]: five upgradient wells (MW-5, MW-9, MW-18, MW-0902 and GLLE-08) and 14 downgradient wells along the landfill and impoundments next to the New River (Fig. 1). In total, 21 constituents were monitored: 19 metals and metalloids, sulfate and cyanide. Eight constituents exceeded applicable groundwater guidelines in at least one well: arsenic (As), boron (B), cobalt (Co), lead (Pb), manganese (Mn), molybdenum (Mo), strontium (Sr) and sulfate ($SO_4^{2-}$). The remaining thirteen stayed below guidelines: antimony, barium, beryllium, cadmium, chromium, copper, cyanide, mercury, nickel, selenium, silver, thallium and uranium. In total, 17 of the 19 wells (89%) recorded at least one exceedance [44].

The exceeding constituents are consistent with a CCR leachate source. Boron, strontium, molybdenum and sulfate are soluble components of coal ash and are used as tracers of ash-derived water [2]. As and Mo form oxyanions whose release depends on pH and redox conditions [1], and Mn and Co are mobilized mainly under reducing conditions, which can develop in saturated ash and in the underlying alluvium [2]. Because several constituents exceeded guidelines together and their mobilities differ, no single constituent describes groundwater contamination conditions at the site, which motivates a composite index.

# IV. Data and Methods

## *A. Hazard Reference*

Groundwater data were processed with the HQ–HI method of the preceding study [11]. Temporally aggregated concentrations were standardized by constituent-specific reference thresholds:

$$HQ_{i,j} = \frac{C_{i,j}}{T_i} \tag{1}$$

where $C_{i,j}$ is the aggregated concentration of constituent $i$ at well $j$ and $T_i$ is its reference threshold (e.g., the USEPA maximum contaminant level) [11]. The HQs were summed into a composite HI:

$$HI_j = \sum_{i=1}^{n} H\,Q_{i,j} \tag{2}$$

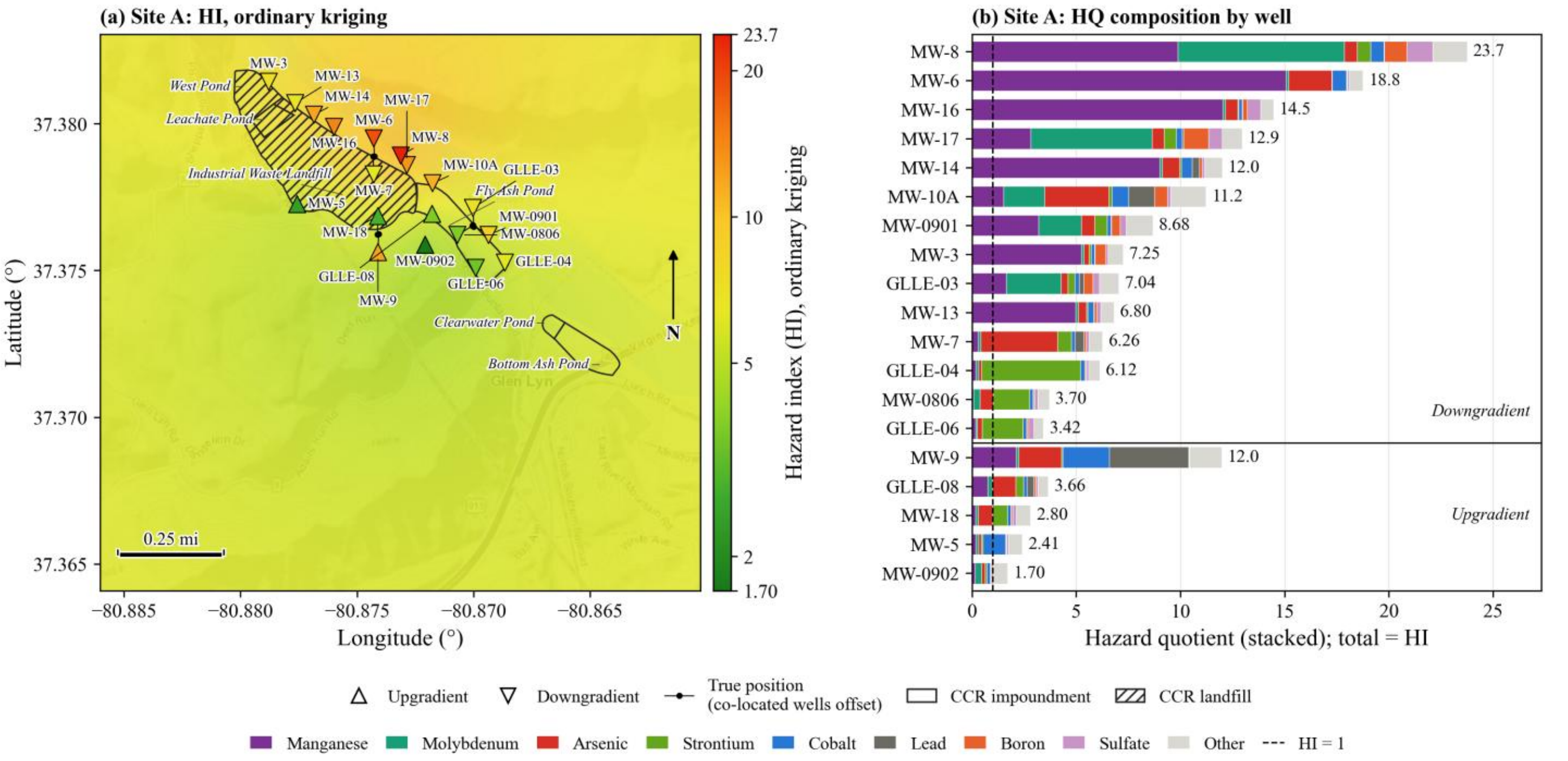

Fig. 2. (a) Composite hazard index (HI) at Site A interpolated by ordinary kriging, with the monitoring wells colored by their HI (logarithmic color scale; △ upgradient, ▽ downgradient). (b) Hazard-quotient (HQ) composition of the HI at each well, grouped into downgradient and upgradient wells; the bars stack the HQs of the constituents, and the dashed line marks HI = 1.

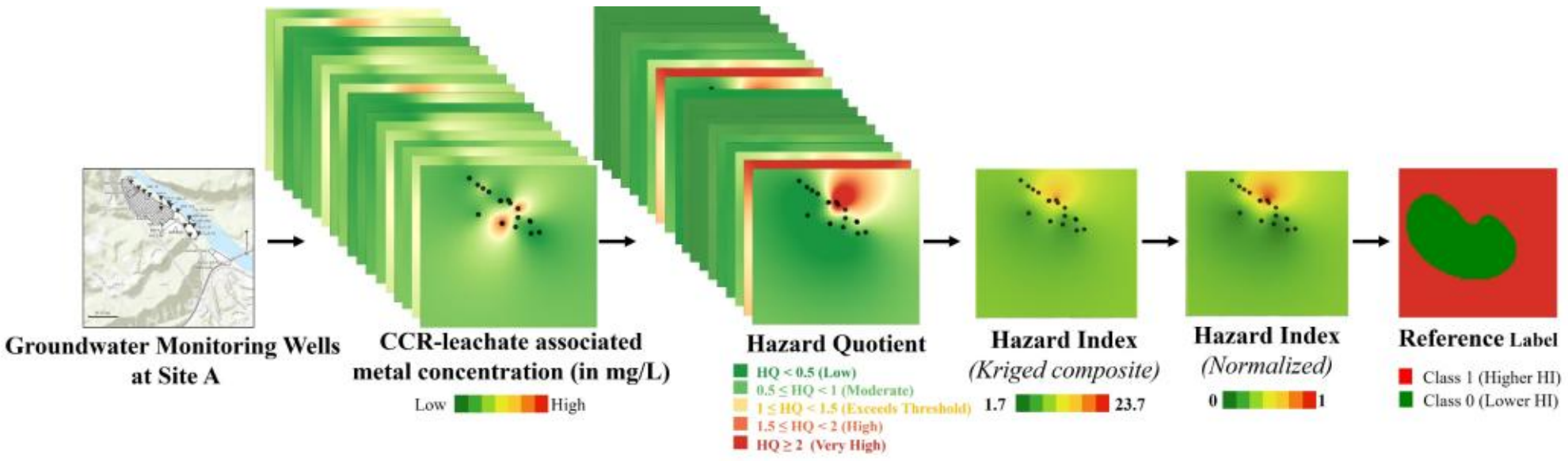

Fig. 3. Workflow for the hazard reference: groundwater monitoring wells at Site A; constituent concentrations; hazard quotients (HQ) relative to the reference thresholds; kriged composite HI; normalized HI; and binary reference labels (class 0, lower HI; class 1, higher HI).

where $n$ is the number of constituents. The HI exceeded 1 at every well, from 1.70 at the upgradient well MW-0902 to 23.7 at the downgradient well MW-8 (Fig. 2). The highest values occurred in the downgradient wells along the northern margin of the industrial landfill next to the New River (MW-8, 23.7; MW-6, 18.8; MW-16, 14.5; MW-17, 12.9). The constituents that dominated the HI differed between wells: manganese at MW-6, MW-16, MW-14, MW-13 and MW-3; molybdenum at MW-17 and GLLE-03 and, together with manganese, at MW-8; arsenic at MW-7 and MW-10A; and strontium at GLLE-04, MW-0806 and GLLE-06. The upgradient well MW-9 also had a high HI (12.0), contributed mainly by lead, cobalt, arsenic and manganese, so a high HI is not restricted to downgradient positions. No single constituent therefore describes the spatial pattern, which supports the use of the composite index.

The well-level HI values were interpolated by ordinary kriging (Fig. 3); the kriged surface ranged from 1.7 to 23.7 and was min–max normalized:

$$HI_{\mathrm{norm}}(x) = \frac{HI(x) - HI_{\min}}{HI_{\max} - HI_{\min}} \tag{3}$$

The normalized surface was converted into binary reference classes:

$$y(x) = \begin{cases} 1, & HI_{\mathrm{norm}}(x) \geq \alpha \\ 0, & HI_{\mathrm{norm}}(x) < \alpha \end{cases} \tag{4}$$

where class 1 denotes higher and class 0 lower relative HI, and $\alpha$ is the 25th percentile of $HI_{\mathrm{norm}}$ over the study area. A relative threshold was required because the kriged HI exceeded 1 everywhere (minimum 1.7), so the conventional threshold HI = 1 would assign every pixel to one class. Because the HI is normalized to the site, its values are meaningful relative to one another, and the 25th percentile isolates the quarter of the area with the lowest relative burden. The resulting classes are imbalanced (approximately 25% class 0 and 75% class 1), so Cohen's κ and class-wise metrics were reported together with accuracy. The labels denote relative rather than absolute hazard.

### *B. Multisource Geospatial Feature Stack*

A 66-band multisource stack was assembled on a 10-m grid of 212 × 288 pixels. Sentinel-2 surface reflectance comprised B1–B8, B8A, B9, B11 and B12. Vegetation and biophysical indices comprised NDVI, MTCI, SAVI, CIRE, NDRE1, NDRE2, LAI, LCC and CCC. Temporal information was represented by the coefficient of spatio-temporal variation (CSTV) and the rate of change (RoC) of the nine vegetation and biophysical index variables. Ancillary predictors were Dynamic World land-cover probabilities [45]; elevation, slope, aspect and curvature [46]; soil moisture, pH and sand, silt and clay fractions [47]; temperature and humidity [48]; and summer precipitation, upstream contributing area, height above nearest drainage (HAND) and topographic wetness index (TWI) [49].

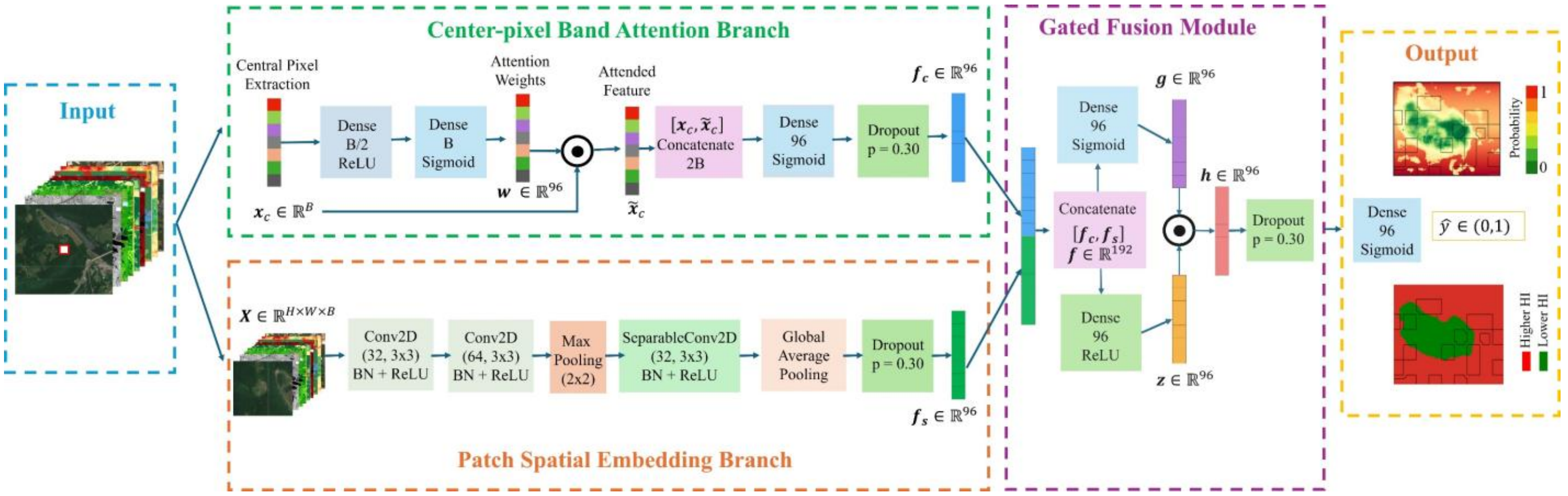

Fig. 4. Architecture of HazardFuse. The center-pixel branch weights the 63 bands of the target pixel by band attention and embeds the raw and attended vectors; the spatial branch encodes the 21 × 21 × 63 patch by two 3 × 3 convolutions with batch normalization, max pooling, a depthwise-separable convolution and global average pooling; a sigmoid gate modulates the joint embedding before the sigmoid output. The model has 97,121 parameters.

The stack was cleaned before modeling. The Sentinel-2 QA60 band and the categorical Dynamic World label band were removed; bands in which one value covered at least 99% of the valid pixels were removed (soil moisture and summer precipitation, whose coarse source resolution makes them nearly constant over the site); and aspect was replaced by its sine and cosine. The resulting 63 bands (TABLE I) were standardized with the mean and standard deviation of the training-block pixels. The bands represent heterogeneous predictors rather than a spectral continuum, which motivates the band-attention branch of HazardFuse.

TABLE I
FEATURE GROUPS OF THE 63-BAND INPUT STACK

| Feature group | Bands | Variables |
|---|---|---|
| Sentinel-2 reflectance | 12 | B1, B2, B3, B4, B5, B6, B7, B8, B8A, B9, B11, B12 |
| Vegetation indices | 9 | NDVI, MTCI, SAVI, CIRE, NDRE1, NDRE2, LAI, LCC, CCC |
| CSTV | 9 | CSTV of the nine vegetation variables |
| Rate of change (RoC) | 9 | RoC of the nine vegetation variables |
| Land cover | 10 | Dynamic World probabilities: bare, built, crops, flooded vegetation, grass, shrubs, snow, trees, vegetation, water |
| Topographic | 5 | DEM, slope, sine and cosine of aspect, curvature |
| Soil | 4 | pH; sand, silt and clay fractions (0–5 cm) |
| Climate | 2 | Temperature, humidity |
| Hydrologic | 3 | Upstream contributing area, HAND, TWI |
| Total | 63 | |

Removed before modeling: Sentinel-2 QA60, the categorical Dynamic World label band, and soil moisture and summer precipitation (one value on ≥99% of the valid pixels); aspect was replaced by its sine and cosine.

### C. Spatial Partitioning

The raster contained 61,056 valid pixels, 15,912 in class 0 and 45,144 in class 1 (class-1 fraction 0.739). To limit spatial leakage, the domain was divided into 32 × 32-pixel blocks, and whole blocks were assigned to training, validation and test subsets with target shares of 0.70, 0.15 and 0.15. Of 500 random assignments, the one closest to the target shares and to the site-wide class balance was retained. Training pixels within 8 pixels of a validation or test block, and validation pixels within 8 pixels of a test block, were excluded as a buffer (TABLE II). The validation blocks were used only for early stopping and learning-rate control; all comparisons are reported on the test blocks.

TABLE II
SPATIAL PARTITION OF THE 61,056 VALID PIXELS

| Partition | Pixels | Share (%) | Class-1 fraction |
|---|---|---|---|
| Training | 30,828 | 50.49 | 0.756 |
| Buffer (excluded) | 13,555 | 22.20 | 0.687 |
| Validation | 7,457 | 12.21 | 0.761 |
| Test | 9,216 | 15.09 | 0.743 |
| Total | 61,056 | 100.00 | 0.739 |

### D. HazardFuse Architecture

For each target pixel, HazardFuse receives a patch $\mathbf{X} \in \mathbb{R}^{P \times P \times B}$ centered on it, with $P = 21$ and $B = 63$; the raster is reflect-padded so that every valid pixel has a full patch. The network has a center-pixel band-attention branch, a patch spatial-embedding branch and a gated fusion module (Fig. 4). All sizes given below are those of the fitted network.

#### 1) Band-Attention Branch

The center-pixel vector $\mathbf{x}_c \in \mathbb{R}^B$ is cropped from the patch. Band-attention weights are computed by two dense layers,

$$\mathbf{a} = \sigma(\mathbf{W}_2\, \delta(\mathbf{W}_1 \mathbf{x}_c + \mathbf{b}_1) + \mathbf{b}_2) \tag{5}$$

where $\delta$ is the ReLU, $\sigma$ the sigmoid, $\mathbf{W}_1$ has $\max(16, \lfloor B/2 \rfloor) = 31$ rows and $\mathbf{a} \in [0,1]^B$. The attended vector $\tilde{\mathbf{x}}_c = \mathbf{x}_c \odot \mathbf{a}$ is concatenated with the original vector, so that the original information is preserved, and projected to a 96-dimensional embedding:

$$\mathbf{f}_c = \mathrm{Drop}_{0.3}\big(\delta(\mathbf{W}_c[\mathbf{x}_c, \tilde{\mathbf{x}}_c] + \mathbf{b}_c)\big) \tag{6}$$

#### 2) Spatial-Embedding Branch

The patch is encoded by two 3 × 3 convolutions with 32 and 64 filters, each followed by batch normalization [50] and ReLU, a 2 × 2 max pooling, a 3 × 3 depthwise-separable convolution with 96 filters and ReLU [36] , and global average pooling (GAP):

$$\mathbf{f}_s = \mathrm{Drop}_{0.3}\Big(\mathrm{GAP}\Big(\delta\Big(\mathrm{SepConv}\big(\mathrm{MaxPool}(\mathbf{F}_2)\big)\Big)\Big)\Big) \tag{7}$$

where $\mathbf{F}_2 \in \mathbb{R}^{21\times21\times64}$ is the output of the second convolution and $\mathbf{f}_s \in \mathbb{R}^{96}$.

### 3) Gated Fusion and Output

The two embeddings are concatenated, $\mathbf{f} = [\mathbf{f}_c, \mathbf{f}_s] \in \mathbb{R}^{192}$, and transformed in parallel into a joint representation and a gate,

$$\mathbf{z} = \delta(\mathbf{W}_z\mathbf{f} + \mathbf{b}_z), \quad \mathbf{g} = \sigma\left(\mathbf{W}_g\mathbf{f} + \mathbf{b}_g\right) \tag{8}$$

with $\mathbf{z}, \mathbf{g} \in \mathbb{R}^{96}$. The gated representation and the class-1 probability are

$$\hat{y} = \sigma(\mathbf{w}_o^{\top}\mathrm{Drop}_{0.3}(\mathbf{z} \odot \mathbf{g}) + b_o) \tag{9}$$

The gate lets the network rely more on the target-location bands or on the neighborhood context, depending on the sample.

## E. Training, Calibration and Mapping

HazardFuse was trained with the class-weighted binary cross-entropy

$$\mathcal{L} = -\frac{1}{N}\sum_{i=1}^{N} w_{y_i}\,\ell_i \tag{10}$$

$$\ell_i = y_i\log\hat{y}_i + (1 - y_i)\log(1 - \hat{y}_i) \tag{11}$$

with $w_k = N/(2N_k)$, where $N_k$ is the number of training pixels of class $k$ (here $w_0 = 2.05$ and $w_1 = 0.66$). Training used Adam [51] with a learning rate of $10^{-3}$, mini-batches of 256 patches, random 90° rotations and horizontal and vertical flips of each training patch, at most 100 epochs, halving of the learning rate after 6 epochs without improvement of the validation loss (minimum $10^{-6}$), and early stopping after 20 such epochs with the best weights restored. The procedure is summarized in Algorithm 1. Every model was trained three times on the same split, with random seeds 42, 43 and 44, which change the weight initialization and the random sampling during training.

Raw scores of different models are not on a common probability scale (for example, tree votes and sigmoid outputs), and a fixed threshold of 0.5 on raw scores would compare models at different operating points. The scores $s(x)$ of every model were therefore calibrated with one rank-based rule that uses only the training labels. With $r(x) \in [0,1]$ the rank of $s(x)$ among the valid pixels divided by their number minus one, and $\pi$ the class-1 fraction of the training pixels ($\pi = 0.756$), the cut is $\tau = 1 - \pi$ and

$$\hat{p}(x) = \begin{cases} 0.5\,r(x)/\tau, & r(x) < \tau \\ 0.5 + 0.5\,\dfrac{r(x) - \tau}{1 - \tau}, & r(x) \geq \tau \end{cases} \tag{12}$$

A pixel is assigned to class 1 where $\hat{p}(x) \geq 0.5$. Every model therefore assigns the same share of the site to class 1, equal to the training prevalence, and differs only in where it places the higher-HI class. The calibration preserves the ranking of the scores and uses no information from the validation or test blocks.

**Algorithm 1:** HazardFuse training and hazard mapping

**Input:** band stack $X \in \mathbb{R}^{H\times W\times B}$ ($B = 63$), reference labels $y$, block size 32 px, buffer 8 px, patch size , seeds 42, 43 and 44 (one run each)$P = 21$
**Output:** fitted network $f_\theta$, calibrated class-1 map $\hat{p}$, class map $\hat{y}$

```
Split the raster into 32 × 32-px blocks; assign blocks to training,
  validation and test (0.70/0.15/0.15) by the best of 500 random
  assignments; exclude an 8-px buffer
Standardize every band with the mean and SD of the training pixels;
  reflect-pad X by ⌊P/2⌋
Class weights w_k ← N/(2N_k) from the training pixels
for epoch = 1, …,100 do
    for each mini-batch of 256 training pixels do
        Extract P × P × B patches; apply a random 90° rotation and
          flips
        a ← σ(W_2 δ(W_1 x_c)); f_c ← Drop(δ(W_c[x_c, x_c ⊙ a]))  ▷ band
          attention
        f_s ←
          Drop(GAP(δ(SepConv(MaxPool(CBR_64(CBR_32(X)))))))
          ▷ spatial embedding
        ŷ ← σ(w_o^T Drop(δ(W_z f) ⊙ σ(W_g f))), f = [f_c, f_s]  ▷ gated
          fusion
        Update θ by Adam on the class-weighted cross-entropy (10)
    end for
    Validation loss: halve the learning rate after 6 epochs without
      improvement; keep the best weights; stop after 20 epochs without
      improvement
end for
s(x) ← f_θ(patch centered on x) for every valid pixel x
r(x) ← rank(s(x))/(N_v − 1); τ ← 1 − π (π: class-1 fraction of the
  training pixels)
p̂(x) ← calibrated score by (12); ŷ(x) ← 1 if p̂(x) ≥ 0.5, else 0
return f_θ, p̂, ŷ
```

## F. Benchmark Models

HazardFuse was compared with ten benchmarks from four families (TABLE III). The first family comprises random forest [18] and XGBoost [52], the standard pixel-wise approach for heterogeneous predictors [19]. Each pixel was represented by the 63 static bands and the 59 non-context bands of each of the seven yearly Sentinel-2 stacks (2019–2025), 476 features in total. Random forest used 300 trees, a minimum leaf size of 5, $\sqrt{p}$ features per split and balanced class weights; XGBoost used 600 trees of depth 6, a learning rate of 0.05, row and column subsampling of 0.8, a minimum child weight of 5 and a positive-class weight equal to the class ratio. Neither was tuned on the validation blocks.

The second family is a patch CNN [20], which classifies the center pixel of a 21 × 21 patch with two convolution stages (32 and 64 filters, batch normalization, ReLU and max pooling), global average pooling and a dense layer of 128 units with dropout of 0.5. It shares the center-pixel formulation of HazardFuse but has no band attention or gated fusion, so it is the most direct deep-learning baseline. The third family comprises five encoder–decoder networks derived from one base architecture: U-Net [28], U-Net++ [29], ResU-Net [31], SE-ResU-Net [13], [31] and Attention U-Net [53]. They differ by one modification at a time, so their comparison isolates dense skip pathways, residual learning, channel attention on learned feature maps and spatial attention. The fourth family comprises DeepLabv3+ [33], with atrous spatial pyramid pooling (rates 1, 2 and 4) and image pooling at output stride 4, and SegFormer [34], with a three-stage hierarchical Transformer encoder (widths 32, 64 and 128) and a multilayer-perceptron decoder.

TABLE III
MODELS COMPARED IN THIS STUDY

| Model | Main characteristic | Parameters |
|---|---|---|
| *Pixel-wise machine learning* | | |
| Random forest | Bagged decision trees on 476 pixel features | — |
| XGBoost | Gradient-boosted decision trees on 476 pixel features | — |
| *Patch-based deep learning* | | |
| Patch CNN | Convolution on the 21 × 21 patch, center-pixel output | 91,489 |
| *Encoder–decoder segmentation* | | |
| U-Net | Encoder–decoder with skip connections | 1,944,289 |
| U-Net++ | Nested, densely connected skip pathways | 2,251,777 |
| ResU-Net | U-Net with residual units | 2,033,729 |
| SE-ResU-Net | Residual U-Net with squeeze-and-excitation | 2,061,657 |
| Attention U-Net | Attention gates on the skip connections | 1,977,114 |
| *Multiscale and global context* | | |
| DeepLabv3+ | Atrous spatial pyramid pooling and decoder | 1,513,857 |
| SegFormer | Hierarchical Transformer encoder, MLP decoder | 522,049 |
| *Proposed* | | |
| **HazardFuse** | Center-pixel band attention, patch spatial embedding, gated fusion | 97,121 |

Pixel features: the 63 static bands and the 59 non-context bands of seven yearly stacks (2019–2025). Parameters include batch-normalization statistics.

All models used the same standardized bands (the pixel models also the yearly bands), the same labels, the same spatial split and the same calibration. The deep-learning benchmarks were implemented in PyTorch with a base width of 32, AdamW [54] with a learning rate of $10^{-3}$ and weight decay of $10^{-4}$, class-weighted binary cross-entropy, the same rotations and flips, gradient clipping at a norm of 5, at most 100 epochs, learning-rate halving after 6 and early stopping after 20 epochs without improvement of the validation loss. The segmentation networks were trained on 32 × 32-pixel tiles centered on randomly drawn training pixels (800 tiles per epoch, batches of 32), with the loss computed only at the labeled training pixels of each tile; maps were predicted with overlapping tiles at a stride of 16 pixels and Hann-weighted averaging and evaluated at the same test pixels as HazardFuse.

### G. Evaluation

Predictions were compared with the reference classes at the test pixels ($N$ = 9,216; 2,368 in class 0 and 6,848 in class 1). From the numbers of true positives (TP), true negatives (TN), false positives (FP) and false negatives (FN), with class 1 as the positive class, the overall accuracy and the class-wise precision and recall are

$$\text{Acc} = \frac{TP + TN}{N} \tag{13}$$

$$P_1 = \frac{TP}{TP + FP}, \quad R_1 = \frac{TP}{TP + FN} \tag{14}$$

and $P_0 = TN/(TN + FN)$ and $R_0 = TN/(TN + FP)$ for class 0. The class-wise F1 is $F1_k = 2P_kR_k/(P_k + R_k)$. Cohen's κ [55] corrects the accuracy for the chance agreement $p_e$:

$$\kappa = \frac{\text{Acc} - p_e}{1 - p_e} \tag{15}$$

$$p_e = \frac{(TP + FP)(TP + FN) + (TN + FN)(TN + FP)}{N^2} \tag{16}$$

Because test pixels are spatially autocorrelated, pixel-level standard errors understate uncertainty [17], [37]. Uncertainty was therefore estimated with a paired block bootstrap [56]: the 32 × 32-pixel test blocks were resampled with replacement 2,000 times, all models were scored on the same resampled blocks, and the median and 2.5th and 97.5th percentiles of each score and of each difference from HazardFuse are reported, together with the share of resamples in which the benchmark scored lower. Scores on the training, validation and test blocks were compared to assess overfitting. Test and held-out scores are reported as the mean and sample standard deviation (SD) over the three training runs; the confusion matrices, the comparison across splits, the block bootstrap, the maps and the ablations refer to the run with seed 42.

### H. Ablation and Band Attention

Architectural ablation retrained HazardFuse with one component removed at a time: band attention (the center vector enters the dense embedding directly); the center-pixel branch (spatial branch only); the fusion gate (concatenation and a dense layer); and the spatial-embedding branch (center-pixel branch only). Feature-group ablation retrained the full architecture with one of the nine feature groups of TABLE I removed. All variants used the same split, protocol, seed (42) and calibration. The learned band-attention weights **a** were averaged over the test pixels to describe how strongly each band is passed to the center-pixel embedding.

## V. RESULTS

### A. Overall Performance

Over three training runs (seeds 42, 43 and 44), HazardFuse achieved the highest mean accuracy (0.973 ± 0.002) and κ (0.929 ± 0.005) of the eleven models on the test blocks (TABLE IV). The strongest benchmarks were SegFormer (κ 0.874 ± 0.025), the patch CNN (0.874 ± 0.023) and DeepLabv3+ (0.867 ± 0.053), followed by SE-ResU-Net (0.850 ± 0.073) and ResU-Net (0.848 ± 0.036); the pixel-wise XGBoost (0.823 ± 0.007) and random forest (0.819 ± 0.004) came next, and U-Net++, U-Net and Attention U-Net ranked lowest (mean κ 0.814–0.781). HazardFuse also had the highest mean precision, recall and F1 of both classes: 0.951, 0.943 and 0.947 for class 0 and 0.980, 0.983 and 0.982 for class 1.

HazardFuse ranked first in each of the three runs (TABLE V), with test κ of 0.926, 0.926 and 0.935 and 249 ± 19 misclassified test pixels. Its κ varied little between runs (SD 0.005), as did that of the two tree ensembles (0.004 and 0.007), whereas the SD of the deep-learning benchmarks ranged from 0.023 (patch CNN) to 0.079 (U-Net++). The order of the benchmarks therefore depended on the run: the patch CNN ranked second with seed 42 but sixth and fifth with seeds 43 and 44, and DeepLabv3+ ranked eighth with seed 42 and second with seeds 43 and 44. In every run, HazardFuse made fewer errors than the benchmark with the fewest errors in that run, by 25% (patch CNN, seed 42), 22% (SegFormer, seed 43) and 39% (DeepLabv3+, seed 44). Over the held-out pixels (validation and test blocks together), HazardFuse also had the highest mean κ (0.854 ± 0.007), followed by SegFormer (0.845 ± 0.020) and the patch CNN (0.816 ± 0.022).

TABLE IV
TEST-BLOCK PERFORMANCE OF HAZARDFUSE AND THE BENCHMARKS OVER THREE TRAINING RUNS

| | | | Class 0 (lower HI) | | | Class 1 (higher HI) | | |
|---|---|---|---|---|---|---|---|---|
| Model | Acc. | κ | P | R | F1 | P | R | F1 |
| **HazardFuse** | **0.973** ±0.002 | **0.929** ±0.005 | **0.951** ±0.012 | **0.943** ±0.012 | **0.947** ±0.004 | **0.980** ±0.004 | **0.983** ±0.005 | **0.982** ±0.001 |
| SegFormer | 0.953 ±0.010 | 0.874 ±0.025 | 0.938 ±0.035 | 0.875 ±0.013 | 0.905 ±0.018 | 0.958 ±0.004 | 0.980 ±0.012 | 0.969 ±0.007 |
| Patch CNN | 0.952 ±0.009 | 0.874 ±0.023 | 0.907 ±0.042 | 0.907 ±0.032 | 0.906 ±0.017 | 0.968 ±0.010 | 0.967 ±0.017 | 0.967 ±0.007 |
| DeepLabv3+ | 0.950 ±0.020 | 0.867 ±0.053 | 0.929 ±0.046 | 0.873 ±0.035 | 0.900 ±0.039 | 0.957 ±0.012 | 0.977 ±0.015 | 0.967 ±0.013 |
| SE-ResU-Net | 0.942 ±0.030 | 0.850 ±0.073 | 0.895 ±0.089 | 0.886 ±0.031 | 0.889 ±0.052 | 0.961 ±0.011 | 0.962 ±0.034 | 0.961 ±0.021 |
| ResU-Net | 0.943 ±0.012 | 0.848 ±0.036 | 0.909 ±0.003 | 0.865 ±0.055 | 0.886 ±0.028 | 0.954 ±0.018 | 0.970 ±0.003 | 0.962 ±0.008 |
| XGBoost | 0.934 ±0.003 | 0.823 ±0.007 | 0.906 ±0.008 | 0.831 ±0.006 | 0.867 ±0.005 | 0.943 ±0.002 | 0.970 ±0.003 | 0.956 ±0.002 |
| Random forest | 0.931 ±0.001 | 0.819 ±0.004 | 0.860 ±0.002 | 0.872 ±0.005 | 0.866 ±0.003 | 0.956 ±0.002 | 0.951 ±0.000 | 0.953 ±0.001 |
| U-Net++ | 0.929 ±0.033 | 0.814 ±0.079 | 0.884 ±0.100 | 0.843 ±0.020 | 0.861 ±0.056 | 0.946 ±0.008 | 0.959 ±0.039 | 0.952 ±0.023 |
| U-Net | 0.924 ±0.019 | 0.795 ±0.052 | 0.881 ±0.055 | 0.817 ±0.067 | 0.846 ±0.041 | 0.938 ±0.021 | 0.961 ±0.021 | 0.949 ±0.012 |
| Attention U-Net | 0.916 ±0.023 | 0.781 ±0.053 | 0.841 ±0.084 | 0.841 ±0.064 | 0.838 ±0.038 | 0.946 ±0.019 | 0.941 ±0.039 | 0.943 ±0.017 |

N = 9,216 test pixels (2,368 class 0, 6,848 class 1). Mean (upper line) and standard deviation (lower line) over three training runs with seeds 42, 43 and 44. Acc., accuracy; κ, Cohen's kappa; P, precision; R, recall. Models ordered by mean κ; best mean of each column in bold.

TABLE V
TEST Κ OF EACH TRAINING RUN AND MEAN HELD-OUT Κ

| | Test κ by training seed | | | Test κ | Held-out κ |
|---|---|---|---|---|---|
| Model | 42 | 43 | 44 | Mean ± SD | Mean ± SD |
| **HazardFuse** | **0.926** | **0.926** | **0.935** | **0.929 ± 0.005** | **0.854 ± 0.007** |
| SegFormer | 0.859 | 0.903 | 0.860 | 0.874 ± 0.025 | 0.845 ± 0.020 |
| Patch CNN | 0.900 | 0.863 | 0.858 | 0.874 ± 0.023 | 0.816 ± 0.022 |
| DeepLabv3+ | 0.807 | 0.903 | 0.892 | 0.867 ± 0.053 | 0.802 ± 0.035 |
| SE-ResU-Net | 0.890 | 0.894 | 0.766 | 0.850 ± 0.073 | 0.757 ± 0.097 |
| ResU-Net | 0.867 | 0.871 | 0.807 | 0.848 ± 0.036 | 0.790 ± 0.026 |
| XGBoost | 0.824 | 0.830 | 0.816 | 0.823 ± 0.007 | 0.761 ± 0.007 |
| Random forest | 0.815 | 0.822 | 0.821 | 0.819 ± 0.004 | 0.782 ± 0.002 |
| U-Net++ | 0.723 | 0.852 | 0.867 | 0.814 ± 0.079 | 0.761 ± 0.036 |
| U-Net | 0.739 | 0.805 | 0.842 | 0.795 ± 0.052 | 0.736 ± 0.045 |
| Attention U-Net | 0.777 | 0.730 | 0.836 | 0.781 ± 0.053 | 0.695 ± 0.046 |

Three training runs per model (seeds 42, 43 and 44) with the same split, protocol and calibration. Test: N = 9,216; held-out: validation and test blocks, N = 16,673. SD, sample standard deviation over the three runs. Models ordered by mean test κ; best value of each column in bold.

The confusion matrices of the run with seed 42 (TABLE VI) show that the models differed mainly in errors on the lower-HI class. HazardFuse misclassified 261 test pixels (2.8%): 110 false positives (class 0 predicted as class 1) and 151 false negatives. The patch CNN made 346 errors, so in this run HazardFuse made 25% fewer errors than the strongest benchmark, and 60% fewer than random forest (655). Eight of the ten benchmarks made more false positives than false negatives, with 225–603 false positives against 110 for HazardFuse. Because the calibration fixes the site-wide share of class 1 for every model, these false positives show that the benchmarks extended the higher-HI class into lower-HI areas, and their class-0 recall was correspondingly lower (0.745–0.905 against 0.954). SE-ResU-Net and the patch CNN reached higher class-1 recall than HazardFuse (0.990 and 0.982 against 0.978) and SE-ResU-Net a higher class-0 precision (0.968 against 0.937), but at the cost of 304 and 225 false positives. Random forest and U-Net++ were the only benchmarks with more false negatives than false positives (340 and 590). HazardFuse was therefore the most balanced model: its precision and recall for the two classes all lay between 0.937 and 0.984.

TABLE VI
TEST-BLOCK CONFUSION MATRICES (SEED 42)

| | Class 0 | | Class 1 | | | |
|---|---|---|---|---|---|---|
| Model | TN | FP | FN | TP | Errors | Error (%) |
| **HazardFuse** | 2,258 | 110 | 151 | 6,697 | **261** | **2.8** |
| Patch CNN | 2,143 | 225 | 121 | 6,727 | 346 | 3.8 |
| SE-ResU-Net | 2,064 | 304 | 69 | 6,779 | 373 | 4.0 |
| ResU-Net | 2,118 | 250 | 215 | 6,633 | 465 | 5.0 |
| SegFormer | 2,094 | 274 | 218 | 6,630 | 492 | 5.3 |
| XGBoost | 1,955 | 413 | 188 | 6,660 | 601 | 6.5 |
| Random forest | 2,053 | 315 | 340 | 6,508 | 655 | 7.1 |
| DeepLabv3+ | 1,976 | 392 | 278 | 6,570 | 670 | 7.3 |
| Attention U-Net | 1,817 | 551 | 196 | 6,652 | 747 | 8.1 |
| U-Net | 1,765 | 603 | 275 | 6,573 | 878 | 9.5 |
| U-Net++ | 1,960 | 408 | 590 | 6,258 | 998 | 10.8 |

Class 0 and class 1: reference classes. TN, class 0 predicted as class 0; FP, class 0 predicted as class 1; FN, class 1 predicted as class 0; TP, class 1 predicted as class 1. N = 9,216 (2,368 class 0, 6,848 class 1). Fewest errors in bold.

### *B. Spatial Predictions*

Fig. 5 shows the class-1 score, the classification and the errors of every model over the whole site. Accuracy was computed only within the test blocks; the whole-site error maps also include the training, buffer and validation areas and show how each model reproduces the reference pattern. All models recovered the lower-HI zone in the center of the site, but they differed in how closely they followed its boundary and in the spatial texture of their scores.

HazardFuse produced a coherent score field, with the lowest scores in the core of the lower-HI zone and the highest along the northern margin of the landfill near the downgradient wells. Its classification followed the reference boundary closely, and its whole-site errors were limited to narrow false-positive slivers along the northern and western edges of the lower-HI zone and short false-negative segments on its eastern edge. The patch CNN produced a similar, smoother pattern with wider false-positive margins. The encoder–decoder networks and DeepLabv3+ placed the class boundary inside the lower-HI zone over longer stretches, which produced broad false-positive bands along its margin; U-Net, U-Net++, Attention U-Net and DeepLabv3+ also produced false-negative bands outside its southern and western edges or isolated lower-HI patches away from it. Their score maps show a regular grid texture from tile-based prediction, most visible for DeepLabv3+, SE-ResU-Net and ResU-Net. Random forest and XGBoost produced speckled score maps with scattered errors inside and around the lower-HI zone, as expected for pixel-wise models without neighborhood information.

One error was shared by all models: in the validation block at the northern margin of the lower-HI zone, a wedge of reference class-0 pixels was assigned to class 1 by every model (held-out error maps in Fig. 5). This is the systematic validation error discussed in Section V-C.

### *C. Generalization Across Splits*

TABLE VII compares the training, validation and test scores of the run with seed 42. Random forest and XGBoost fitted the training pixels almost perfectly (κ 0.999 and 1.000) but reached κ of 0.815 and 0.824 on the test blocks, a gap of more than 0.17 that indicates overfitting to training locations. The gap of HazardFuse was 0.059 (κ 0.986 to 0.926). It was smaller only for SE-ResU-Net (0.032) and SegFormer (0.055), which fitted the training pixels less closely (κ 0.922 and 0.914) and scored lower on the test blocks.

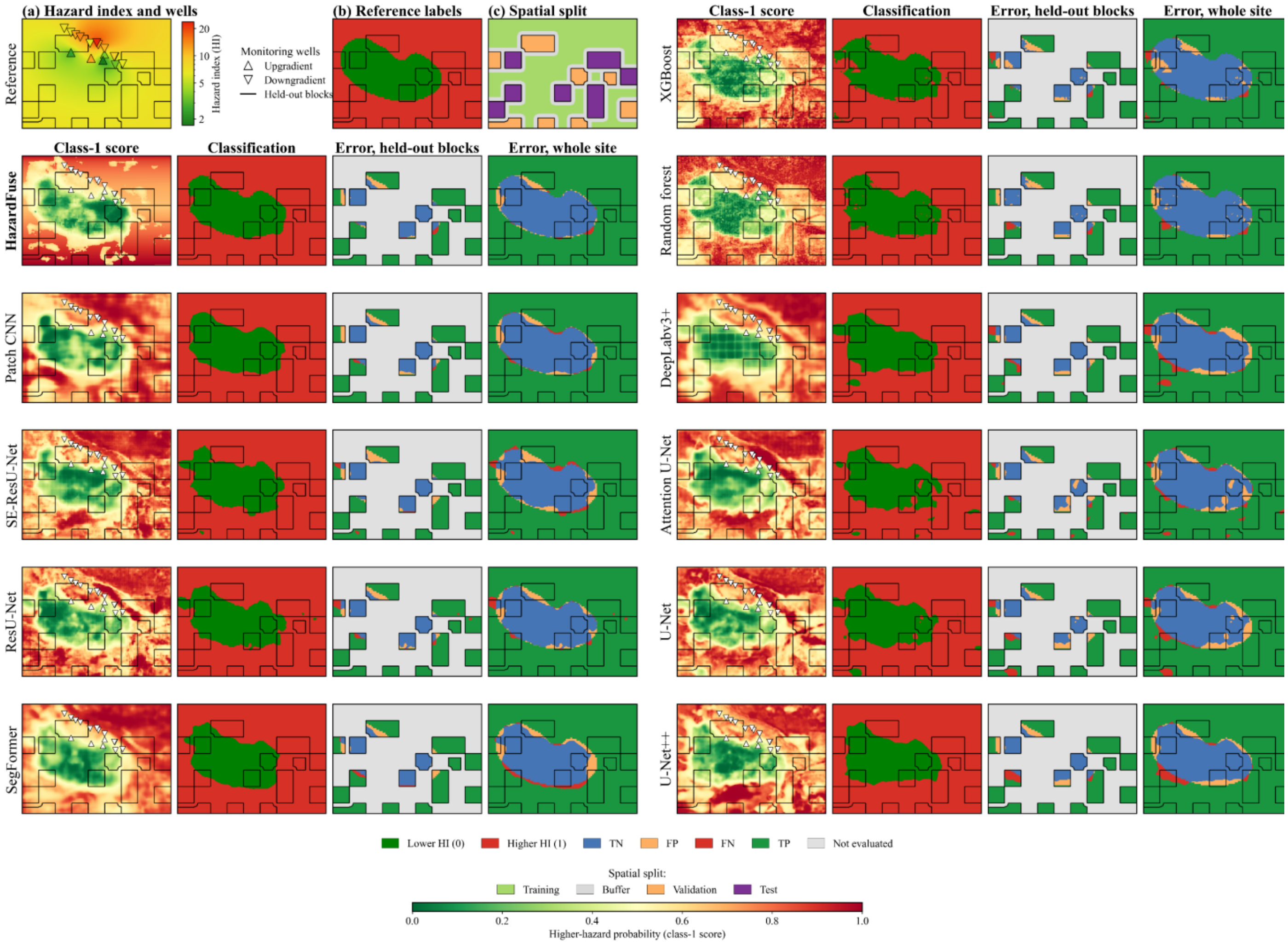

Fig. 5. Spatial predictions of HazardFuse and the ten benchmarks, run with seed 42, ordered by test κ (left: reference row and the first five models; right: the remaining six). Reference row: (a) hazard index with the monitoring wells (△ upgradient, ▽ downgradient), (b) reference labels and (c) spatial split (training, buffer, validation and test). One row per model: class-1 (higher-HI) score, classification, error within the held-out blocks and error over the whole site (TN, FP, FN and TP). Held-out blocks are outlined in black.

TABLE VII
TRAINING, VALIDATION AND TEST SCORES (SEED 42)

| | Accuracy | | | κ | | |
|---|---|---|---|---|---|---|
| Model | Train | Val. | Test | Train | Val. | Test |
| **HazardFuse** | 0.995 | 0.915 | 0.972 | 0.986 | 0.734 | 0.926 |
| Patch CNN | 0.989 | 0.914 | 0.962 | 0.970 | 0.732 | 0.900 |
| SE-ResU-Net | 0.971 | 0.887 | 0.960 | 0.922 | 0.654 | 0.890 |
| ResU-Net | 0.988 | 0.897 | 0.950 | 0.968 | 0.701 | 0.867 |
| SegFormer | 0.968 | 0.926 | 0.947 | 0.914 | 0.774 | 0.859 |
| XGBoost | 1.000 | 0.892 | 0.935 | 1.000 | 0.679 | 0.824 |
| Random forest | 0.999 | 0.914 | 0.929 | 0.999 | 0.733 | 0.815 |
| DeepLabv3+ | 0.957 | 0.909 | 0.927 | 0.880 | 0.743 | 0.807 |
| Attention U-Net | 0.983 | 0.883 | 0.919 | 0.953 | 0.655 | 0.777 |
| U-Net | 0.973 | 0.865 | 0.905 | 0.927 | 0.631 | 0.739 |
| U-Net++ | 0.974 | 0.909 | 0.892 | 0.928 | 0.713 | 0.723 |

All models scored lower on the validation blocks (κ 0.631–0.774) than on the test blocks. On the validation blocks, HazardFuse ranked third (κ 0.734) after SegFormer (0.774) and DeepLabv3+ (0.743), and the top five models were within 0.042 of one another. The lower validation scores came from one systematic error (Fig. 5): every model assigned between 23% and 39% of the validation class-0 pixels to class 1 (class-0 recall 0.612–0.767), and for HazardFuse all 633 validation errors were of this type (class-0 precision and class-1 recall both 1.000). An error shared by all models indicates a part of the validation area where the features resemble higher-HI conditions while the kriged labels assign lower HI, rather than a model-specific weakness. The same pattern held over the three runs: SegFormer had the highest mean validation κ (0.804 ± 0.027) and HazardFuse the second highest (0.745 ± 0.012), and the systematic error on the validation class-0 pixels occurred in every run.

### *D. Block-Bootstrap Uncertainty*

For the run with seed 42, the paired block bootstrap gave HazardFuse a median test κ of 0.922 (95% interval 0.750–0.976; TABLE VIII). The intervals are wide because the test set consists of a limited number of spatially coherent blocks. Because the differences are paired, they are more precise than the individual scores. The 95% interval of the κ difference lay entirely below zero for eight of the ten benchmarks, with median differences from −0.036 (SE-ResU-Net) to −0.212 (U-Net), and the benchmark scored lower in at least 98.6% of resamples. For the patch CNN (median Δκ −0.032; interval −0.072 to +0.010; lower in 91% of resamples) and U-Net++ (−0.211; −0.501 to +0.011; 96%), the interval included zero, so the advantage of HazardFuse over these two models is not established at the 95% level for this set of test blocks. Across the three training runs, however, HazardFuse exceeded both models in every run, by 0.026–0.077 in κ over the patch CNN and 0.068–0.203 over U-Net++.

TABLE VIII
PAIRED BLOCK BOOTSTRAP OF THE TEST BLOCKS

| Model | κ (95% interval) | Δκ (95% interval) | P(Δκ < 0) |
|---|---|---|---|
| **HazardFuse** | 0.922 (0.750, 0.976) | — | — |
| Patch CNN | 0.887 (0.705, 0.975) | −0.032 (−0.072, +0.010) | 0.913 |
| SE-ResU-Net | 0.884 (0.398, 0.970) | −0.036 (−0.367, −0.002) | 0.986 |
| ResU-Net | 0.855 (0.581, 0.942) | −0.065 (−0.186, −0.025) | 0.999 |
| SegFormer | 0.849 (0.208, 0.951) | −0.073 (−0.546, −0.013) | 0.995 |
| XGBoost | 0.813 (0.612, 0.868) | −0.106 (−0.185, −0.045) | 0.996 |
| Random forest | 0.799 (0.470, 0.934) | −0.120 (−0.329, −0.024) | 0.999 |
| DeepLabv3+ | 0.791 (0.452, 0.928) | −0.129 (−0.348, −0.032) | 0.999 |
| Attention U-Net | 0.746 (0.459, 0.935) | −0.166 (−0.301, −0.040) | 0.999 |
| U-Net | 0.703 (0.334, 0.940) | −0.212 (−0.429, −0.022) | 0.999 |
| U-Net++ | 0.716 (0.353, 0.953) | −0.211 (−0.501, +0.011) | 0.963 |

Run with seed 42. Medians and 2.5th–97.5th percentiles over 2,000 resample of the test blocks; Δκ = benchmark − HazardFuse; P(Δκ < 0): share of resamples in which the benchmark scored lower.

### *E. Architectural Ablation*

Every ablated variant scored lower than the full model on accuracy, κ and the F1 of both classes (TABLE IX). Removing the spatial-embedding branch caused by far the largest decline: accuracy decreased by 0.088 (to 0.884), κ by 0.224 (to 0.703), class-0 F1 by 0.164 and class-1 F1 by 0.060, while 26% of the parameters were retained. Neighborhood context therefore carries most of the discriminative information.

Removing band attention caused the second-largest decline (accuracy −0.015, κ −0.042), although the variant retained 90% of the parameters. Its effect was concentrated in class 0: recall decreased from 0.954 to 0.897 and F1 from 0.945 to 0.913, while class-1 recall was unchanged (0.978 and 0.977). Weighting the target-location bands thus mainly helps the network recognize lower-HI pixels. Removing the center-pixel branch reduced κ by 0.022; this variant had a higher class-0 precision (0.962) and class-1 recall (0.988) but a lower class-0 recall (0.896), i.e., without the target-location pathway the network assigned more lower-HI pixels to class 1. Removing the fusion gate caused the smallest decline (accuracy −0.004, κ −0.010) with 81% of the parameters retained. The ranking of the components is spatial branch > band attention > center-pixel branch > fusion gate. Each variant was trained once (seed 42), so differences of about 0.01 in κ, such as that of the fusion gate, should be interpreted with caution.

TABLE IX
ARCHITECTURAL ABLATION OF HAZARDFUSE (TEST BLOCKS)

| | | | Class 0 (lower HI) | | | Class 1 (higher HI) | | |
|---|---|---|---|---|---|---|---|---|
| Configuration | Acc. | κ | P | R | F1 | P | R | F1 |
| **Full model** | 0.972 | 0.926 | 0.937 | 0.954 | 0.945 | 0.984 | 0.978 | 0.981 |
| − Band attention | 0.956 | 0.884 | 0.930 | 0.897 | 0.913 | 0.965 | 0.977 | 0.971 |
| − Center-pixel branch | 0.964 | 0.904 | 0.962 | 0.896 | 0.928 | 0.965 | 0.988 | 0.976 |
| − Fusion gate | 0.968 | 0.916 | 0.931 | 0.944 | 0.938 | 0.981 | 0.976 | 0.978 |
| − Spatial branch | 0.884 | 0.703 | 0.759 | 0.805 | 0.781 | 0.931 | 0.911 | 0.921 |

Each variant removes one component and is retrained with the same split, protocol and seed (42).

### *F. Feature-Group Ablation*

Removing any feature group reduced accuracy and κ (TABLE X). The largest declines followed the removal of the soil group (4 bands; κ −0.097, accuracy −0.038), the Sentinel-2 bands (12; −0.095, −0.033), the vegetation indices (9; −0.085, −0.032) and the rate-of-change metrics (9; −0.078, −0.030). Land cover (10; κ −0.063), climate (2; −0.051) and topography (5; −0.047) produced intermediate declines, and CSTV (9; −0.013) and the hydrologic group (3; −0.008) the smallest.

The groups contributed to different aspects of the classification. Without the Sentinel-2 bands, class-0 recall fell from 0.954 to 0.806 while class-1 recall rose to 0.985, so the network extended the higher-HI class into lower-HI areas; the climate (class-0 recall 0.885) and land-cover (0.876) groups behaved similarly. Without the soil group, class-0 precision (0.852) and class-1 recall (0.946) also decreased, i.e., errors increased in both directions. Without the vegetation indices or rate-of-change metrics, precision and recall of class 0 both fell to 0.88–0.89. The small effects of the CSTV and hydrologic groups are consistent with redundancy: CSTV and rate of change summarize the multi-year behavior of the same nine vegetation variables, and HAND, TWI and upstream area are derived from the same elevation model as the topographic group.

TABLE X
FEATURE-GROUP ABLATION OF HAZARDFUSE (TEST BLOCKS)

| | | | Class 0 (lower HI) | | | Class 1 (higher HI) | | |
|---|---|---|---|---|---|---|---|---|
| Configuration | Acc. | κ | P | R | F1 | P | R | F1 |
| **Full input (63)** | 0.972 | 0.926 | 0.937 | 0.954 | 0.945 | 0.984 | 0.978 | 0.981 |
| − Sentinel-2 (12) | 0.939 | 0.831 | 0.948 | 0.806 | 0.871 | 0.936 | 0.985 | 0.960 |
| − Veg. indices (9) | 0.939 | 0.841 | 0.881 | 0.883 | 0.882 | 0.960 | 0.959 | 0.959 |
| − CSTV (9) | 0.967 | 0.913 | 0.926 | 0.946 | 0.936 | 0.981 | 0.974 | 0.977 |
| − RoC (9) | 0.942 | 0.848 | 0.885 | 0.890 | 0.887 | 0.962 | 0.960 | 0.961 |
| − Topographic (5) | 0.954 | 0.880 | 0.907 | 0.915 | 0.911 | 0.970 | 0.967 | 0.969 |
| − Soil (4) | 0.934 | 0.829 | 0.852 | 0.898 | 0.874 | 0.964 | 0.946 | 0.955 |
| − Climate (2) | 0.953 | 0.875 | 0.929 | 0.885 | 0.907 | 0.961 | 0.977 | 0.969 |
| − Hydrologic (3) | 0.969 | 0.918 | 0.943 | 0.935 | 0.939 | 0.978 | 0.981 | 0.979 |
| − Land cover (10) | 0.948 | 0.863 | 0.920 | 0.876 | 0.897 | 0.958 | 0.974 | 0.966 |

Number of removed bands in parentheses; each variant retrained with the same split, protocol and seed.

### *G. Band-Attention Weights*

The mean attention weights ranged from 0.208 (B9) to 0.791 (CSTV of CIRE) over the 63 bands (Fig. 6). By group, the climate (mean 0.625), land-cover (0.575) and hydrologic (0.553) bands received the highest weights and the topographic bands the lowest (0.395). The most strongly passed individual bands were CSTV of CIRE, the Dynamic World built-area probability (0.788), HAND (0.715), temperature (0.706), the shrub probability (0.702) and the rate of change of SAVI (0.684). The attention ranking differs from the ablation ranking: CSTV of CIRE received the highest weight although removing the whole CSTV group changed κ by only −0.013, and the soil bands received intermediate weights (0.517) although their removal caused the largest decline. Attention describes how strongly each band enters the center-pixel embedding, whereas ablation measures the information that the other bands cannot replace; the two are complementary and neither is a causal attribution.

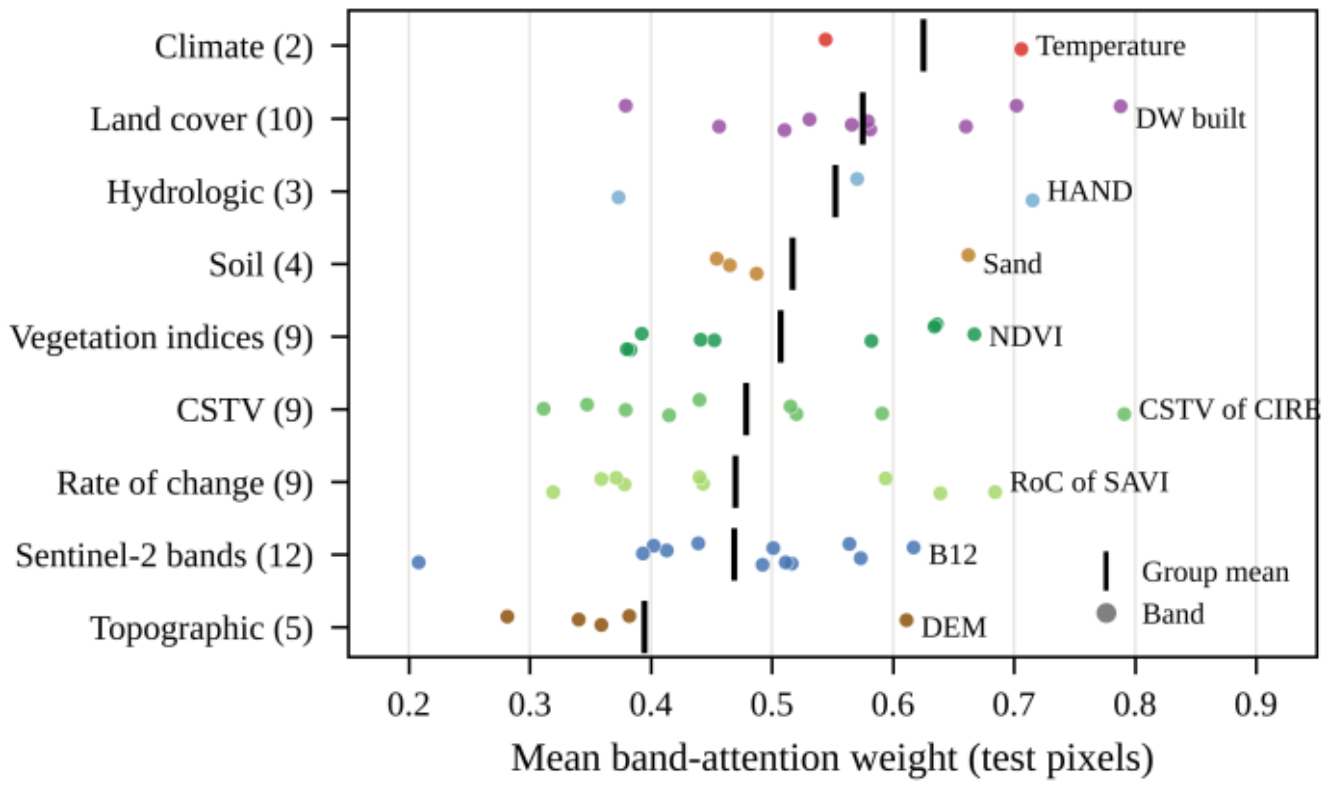

Fig. 6. Mean band-attention weight of each of the 63 input bands over the test pixels, grouped by feature group (number of bands in parentheses). Vertical bars mark group means; the band with the highest weight in each group is named.

## VI. Discussion

### A. Combining Target-Location and Neighborhood Information

The results show that combining the multisource band vector of the target location with its neighborhood context is effective for mapping conditions associated with the composite HI. HazardFuse reached a mean accuracy of 0.973 and a mean κ of 0.929 on the test blocks over three training runs, although about three quarters of the test pixels belonged to the higher-HI class. Its class-0 F1 (0.947) and recall (0.943) show that this performance was not driven by the majority class, and its κ varied little between runs (SD 0.005), so the result does not depend on a favorable initialization. Because all models were calibrated to the same site-wide share of class 1, the comparison reflects where each model places the higher-HI class rather than how its threshold was chosen.

### B. Contribution of the Architectural Components

Removing the spatial branch reduced κ by 0.224, which confirms that the neighborhood of a pixel contains information about HI conditions that its own band vector does not. Band attention was the second most important component (κ −0.042) and acted mainly on the lower-HI class, whose recall fell by 0.056 without it. This supports the premise that heterogeneous predictors should be weighted at the target location before they are combined with the spatial context. The center-pixel pathway contributed κ 0.022, and the fusion gate 0.010; the latter is about twice the run-to-run SD of the full model (0.005) and was measured in a single run. The gate is retained because it adds the sample-dependent weighting of the two streams at moderate cost (19% of the parameters), but its benefit should be confirmed with repeated runs of the ablated variants.

### C. Contribution of the Feature Groups

All feature groups contributed, which differs from the expectation that derived or redundant groups could be removed without loss. Soil, Sentinel-2 reflectance, vegetation indices and their rates of change were the most informative groups, consistent with the premise that groundwater-derived hazard is expressed at the surface mainly through vegetation response, modulated by soil texture and chemistry. The Sentinel-2, climate and land-cover groups mainly helped the network separate lower-HI areas, so that without them the higher-HI class extended into lower-HI areas. The CSTV and hydrologic groups added the least beyond the other groups, which reflects overlap with the rate-of-change and topographic groups rather than an absence of information; CSTV of CIRE received the highest attention weight of all bands. CSTV was retained because it describes a physically distinct property, the spatio-temporal stability of vegetation condition, which has been used to separate persistent metal-related stress from short-term stress [9], [10]. Ablation measures how much the predictions depend on an input given the others; it identifies redundancy among predictors, not the processes that cause the HI pattern.

### D. Comparison With the Benchmarks

HazardFuse had the highest mean accuracy, κ and class-wise F1 of the eleven models, ranked first in each of three training runs and made 22–39% fewer errors than the best benchmark of each run. For the run with seed 42, the paired block bootstrap supports its advantage over eight benchmarks but not over the patch CNN and U-Net++, whose intervals included zero; across the three runs, however, HazardFuse exceeded both in every run. It was also the most stable of the deep-learning models (κ SD 0.005 against 0.023–0.079), whereas the order of the benchmarks changed between runs. The closest competitors were SegFormer and the patch CNN (mean κ 0.874). The patch CNN shares the center-pixel formulation with HazardFuse but has neither band attention nor gated fusion; the difference between them corresponds to the attention and fusion components identified in the ablation. Segmentation networks with hundreds of thousands to millions of parameters did not outperform HazardFuse (97,121 parameters) and at best matched the patch CNN (91,489), which is consistent with the sparse, point-derived nature of the supervision. The pixel-wise tree ensembles fitted the training pixels almost perfectly and generalized worst, as expected for models without neighborhood information that use many correlated predictors.

### E. Limitations

The reference labels were derived by kriging HI values from 19 wells, so uncertainty in well coverage, temporal aggregation, reference thresholds and the variogram propagates into the labels. This uncertainty is greatest far from the wells and near the class boundary, and the error shared by all models on the validation blocks indicates such an area. The HI is normalized to the site and represents relative metal burden; it is neither a probability nor a regulatory risk measure.

All comparisons use one spatial partition. The model comparison was repeated with three training seeds, but the block bootstrap and the ablations were computed for one run (seed 42). The bootstrap quantifies the uncertainty due to the choice of test blocks, and three runs give only a coarse estimate of the variability between runs, so differences of about 0.01 in κ in the ablations should not be interpreted as definitive. The groundwater samples were collected between 2010 and 2014, whereas the satellite data span 2019–2025; the analysis assumes that the spatial pattern of hazard persisted, which site changes such as the capping of the Auxiliary Fly Ash Pond in 2014 may weaken locally [57], [58]. Finally, all results were obtained at one site, so the transferability of the trained model and of the feature rankings to other CCR facilities remains to be established.

## VII. Conclusion

Monitoring wells measure leached-metal hazard only at discrete points. This study investigated whether multisource satellite and geospatial data can extend these measurements into continuous hazard maps, and it addressed the question with HazardFuse, a network that learns the relevance of heterogeneous bands at the target location through band attention, learns neighborhood structure through a spatial-embedding branch and combines the two through a learned gate. Supervised by a composite HI of the constituents measured in 19 monitoring wells and evaluated on spatially separated test blocks at the former Glen Lyn Power Station, HazardFuse reached a mean accuracy of $0.973 \pm 0.002$ and a $\kappa$ of $0.929 \pm 0.005$ over three training runs, with precision, recall and F1 of 0.951, 0.943 and 0.947 for the lower-HI class and 0.980, 0.983 and 0.982 for the higher-HI class. It ranked first among eleven models in every run, made 22–39% fewer errors than the best of ten benchmarks in each run, and varied less between runs than any deep-learning benchmark. Neighborhood context was the most important source of information, followed by band attention, which mainly improved recognition of lower-HI areas. All feature groups contributed, with soil, Sentinel-2 reflectance, vegetation indices and their rates of change the most informative.

The contribution of this work is a reproducible way to translate discrete groundwater monitoring into maps of HI-associated conditions around CCR facilities. It combines a compact model of 97,121 parameters, one documented procedure (**Algorithm 1**) and a calibration that places all models at the same operating point. The probability maps show where a higher relative metal hazard is most likely and where predictions are least certain, which provides a basis for placing additional groundwater sampling. The approach is not specific to CCR. Any problem in which a hazard is measured at sparse points and has a surface expression in vegetation, soil, terrain or hydrology can be addressed with the same steps: point measurements are converted into a composite index, interpolated into reference classes and related to a multisource satellite and geospatial stack. Examples include tailings storage facilities and acid mine drainage at mine sites, municipal and industrial landfills, legacy industrial sites, and regional groundwater constituents such as arsenic or nitrate. The band attention reports which inputs drive the predictions at each new site, and the open-source implementation allows the model to be retrained on local monitoring data or applied to sites that share the same predictor bands. Such applications require labels from local monitoring, and the transferability of a trained model between sites remains to be tested. The maps describe spatial associations with a study-specific composite HI; they are not a direct detection of leachate, dissolved metals or ecological risk.

## Data and code Availability

The multisource remote sensing data used in this study, is available on Zenodo at https://doi.org/10.5281/zenodo.23823223 [59].

The HazardFuse code is available at https://github.com/salmansust/hazardfuse under the MIT license. The repository provides HazardFuse as a Python package with a command-line interface for training the model and mapping new sites.

## Acknowledgment

This work was supported by the U.S. Department of Energy, National Energy Technology Laboratory, under Award DE-FE0032184.

## References

[1] M. Izquierdo and X. Querol, “Leaching behaviour of elements from coal combustion fly ash: An overview,” *Int. J. Coal Geol.*, vol. 94, pp. 54–66, May 2012, doi: 10.1016/j.coal.2011.10.006.

[2] J. S. Harkness, B. Sulkin, and A. Vengosh, “Evidence for Coal Ash Ponds Leaking in the Southeastern United States,” *Environ. Sci. Technol.*, vol. 50, no. 12, pp. 6583–6592, Jun. 2016, doi: 10.1021/acs.est.6b01727.

[3] US EPA, “Hazardous and Solid Waste Management System; Disposal of Coal Combustion Residuals From Electric Utilities,” Apr. 2015.

[4] US EPA, “Legacy Coal Combustion Residuals Surface Impoundments and CCR Management Units Final Rule,” 2024.

[5] Virginia DEQ, “Glen Lyn Power Station,” https://www.deq.virginia.gov/news-info/shortcuts/permits/waste/coal-ash/glen-lyn-power-station. Accessed: Oct. 26, 2026. [Online]. Available: https://www.deq.virginia.gov/news-info/shortcuts/permits/waste/coal-ash/glen-lyn-power-station

[6] Code of Virginia, *§ 10.1-1402.04, “Closure of certain coal combustion residuals units.”* Accessed: Nov. 26, 2026. [Online]. Available: https://law.lis.virginia.gov/vacode/title10.1/chapter14/section10.1-1402.04/

[7] D. B. U. , C. S. , C. O. , F. V. , G. F. , H. B. , I. C. , L. M. F. , B. P. ,P. , M. P. , M. A. , S. F. , S. O. , Drusch M., “SENTINEL-2: ESA’s Optical High-Resolution Mission for GMES Operational Services,” *Remote Sens. Environ.*, vol. 120, pp. 25–36, 2012.

[8] L. Kooistra *et al.*, “Exploring field vegetation reflectance as an indicator of soil contamination in river floodplains,” *Environmental Pollution*, vol. 127, no. 2, pp. 281–290, Jan. 2004, doi: 10.1016/S0269-7491(03)00266-5.

[9] M. Liu, T. Wang, A. K. Skidmore, and X. Liu, “Heavy metal-induced stress in rice crops detected using multi-temporal Sentinel-2 satellite images,” *Science of the Total Environment*, vol. 637–638, pp. 18–29, Oct. 2018, doi: 10.1016/j.scitotenv.2018.04.415.

[10] S. Karim, B. Nojabaei, and H. Foroutan, “Phyto-identification of leached metal contamination using satellite remote sensing: a case study on a coal-fired power plant,” *Environ. Monit. Assess.*, vol. 198, no. 3, p. 263, Feb. 2026, doi: 10.1007/s10661-026-15123-5.

[11] S. Karim, B. Nojabaei, and H. Foroutan, “An Integrated Remote Sensing and Machine Learning Framework for Mapping Hazard Associated with Leached Metals,” 2026, doi: 10.48550/arXiv.2611.32505.

[12] J. Podgorski and M. Berg, “Global threat of arsenic in groundwater,” *Science (1979).*, vol. 368, no. 6493, pp. 845–850, May 2020, doi: 10.1126/science.aba1510.

[13] J. Hu, L. Shen, and G. Sun, “Squeeze-and-Excitation Networks,” in *2018 IEEE/CVF Conference on Computer Vision and Pattern Recognition*, IEEE, Jun. 2018, pp. 7132–7141. doi: 10.1109/CVPR.2018.00745.

[14] S. Woo, J. Park, J.-Y. Lee, and I. S. Kweon, “CBAM: Convolutional Block Attention Module,” 2018, pp. 3–19. doi: 10.1007/978-3-030-01234-2_1.

[15] H. Sun, X. Zheng, X. Lu, and S. Wu, "Spectral–Spatial Attention Network for Hyperspectral Image Classification," *IEEE Transactions on Geoscience and Remote Sensing*, vol. 58, no. 5, pp. 3232–3245, May 2020, doi: 10.1109/TGRS.2019.2951160.

[16] S. K. Roy, S. Manna, T. Song, and L. Bruzzone, "Attention-Based Adaptive Spectral–Spatial Kernel ResNet for Hyperspectral Image Classification," *IEEE Transactions on Geoscience and Remote Sensing*, vol. 59, no. 9, pp. 7831–7843, Sep. 2021, doi: 10.1109/TGRS.2020.3043267.

[17] D. R. Roberts *et al.*, "Cross-validation strategies for data with temporal, spatial, hierarchical, or phylogenetic structure," *Ecography*, vol. 40, no. 8, pp. 913–929, Aug. 2017, doi: 10.1111/ecog.02881.

[18] L. Breiman, "Random Forests," *Mach. Learn.*, vol. 45, no. 1, pp. 5–32, Oct. 2001, doi: 10.1023/A:1010933404324.

[19] M. Belgiu and L. Drăguţ, "Random forest in remote sensing: A review of applications and future directions," *ISPRS Journal of Photogrammetry and Remote Sensing*, vol. 114, pp. 24–31, Apr. 2016, doi: 10.1016/j.isprsjprs.2016.01.011.

[20] A. Sharma, X. Liu, X. Yang, and D. Shi, "A patch-based convolutional neural network for remote sensing image classification," *Neural Networks*, vol. 95, pp. 19–28, Nov. 2017, doi: 10.1016/j.neunet.2017.07.017.

[21] T. Baltrusaitis, C. Ahuja, and L.-P. Morency, "Multimodal Machine Learning: A Survey and Taxonomy," *IEEE Trans. Pattern Anal. Mach. Intell.*, vol. 41, no. 2, pp. 423–443, Feb. 2019, doi: 10.1109/TPAMI.2018.2798607.

[22] Y. Dai, F. Gieseke, S. Oehmcke, Y. Wu, and K. Barnard, "Attentional Feature Fusion," in *2021 IEEE Winter Conference on Applications of Computer Vision (WACV)*, IEEE, Jan. 2021, pp. 3559–3568. doi: 10.1109/WACV48630.2021.00360.

[23] S. Hochreiter and J. Schmidhuber, "Long Short-Term Memory," *Neural Comput.*, vol. 9, no. 8, pp. 1735–1780, Nov. 1997, doi: 10.1162/neco.1997.9.8.1735.

[24] Y. N. Dauphin, A. Fan, M. Auli, and D. Grangier, "Language modeling with gated convolutional networks. ," in *Proc. 34th Int. Conf. Mach. Learn. (ICML)*, PMLR 70, 2017, pp. 933–941. Accessed: Sep. 30, 2026. [Online]. Available: https://proceedings.mlr.press/v70/dauphin17a.html

[25] J. Arevalo, T. Solorio, M. Montes-y-Gómez, and F. A. González, "Gated Multimodal Units for Information Fusion," *ICLR 2017 Workshop*, Feb. 2017, Accessed: Sep. 30, 2026. [Online]. Available: https://arxiv.org/abs/1702.01992

[26] X. Li, H. Zhao, L. Han, Y. Tong, S. Tan, and K. Yang, "Gated Fully Fusion for Semantic Segmentation," *Proceedings of the AAAI Conference on Artificial Intelligence*, vol. 34, no. 07, pp. 11418–11425, Apr. 2020, doi: 10.1609/aaai.v34i07.6805.

[27] D. Hong *et al.*, "More Diverse Means Better: Multimodal Deep Learning Meets Remote-Sensing Imagery Classification," *IEEE Transactions on Geoscience and Remote Sensing*, vol. 59, no. 5, pp. 4340–4354, May 2021, doi: 10.1109/TGRS.2020.3016820.

[28] O. Ronneberger, P. Fischer, and T. Brox, "U-Net: Convolutional Networks for Biomedical Image Segmentation," May 2015.

[29] Z. Zhou, M. M. Rahman Siddiquee, N. Tajbakhsh, and J. Liang, "UNet++: A Nested U-Net Architecture for Medical Image Segmentation," 2018, pp. 3–11. doi: 10.1007/978-3-030-00889-5_1.

[30] F. I. Diakogiannis, F. Waldner, P. Caccetta, and C. Wu, "ResUNet-a: A deep learning framework for semantic segmentation of remotely sensed data," *ISPRS Journal of Photogrammetry and Remote Sensing*, vol. 162, pp. 94–114, Apr. 2020, doi: 10.1016/j.isprsjprs.2020.01.013.

[31] Z. Zhang, Q. Liu, and Y. Wang, "Road Extraction by Deep Residual U-Net," *IEEE Geoscience and Remote Sensing Letters*, vol. 15, no. 5, pp. 749–753, May 2018, doi: 10.1109/LGRS.2018.2802944.

[32] L. Rundo *et al.*, "USE-Net: Incorporating Squeeze-and-Excitation blocks into U-Net for prostate zonal segmentation of multi-institutional MRI datasets," *Neurocomputing*, vol. 365, pp. 31–43, Nov. 2019, doi: 10.1016/j.neucom.2019.07.006.

[33] L.-C. Chen, Y. Zhu, G. Papandreou, F. Schroff, and H. Adam, "Encoder-Decoder with Atrous Separable Convolution for Semantic Image Segmentation," 2018, pp. 833–851. doi: 10.1007/978-3-030-01234-2_49.

[34] E. Xie, W. Wang, Z. Yu, A. Anandkumar, J. M. Alvarez, and P. Luo, "SegFormer: Simple and Efficient Design for Semantic Segmentation with Transformers," Oct. 2021.

[35] US EPA, "Risk Assessment Guidance for Superfund, Volume I: Human Health Evaluation Manual (Part A). EPA/540/1-89/002.," Washington, DC, 1989.

[36] F. Chollet, "Xception: Deep Learning with Depthwise Separable Convolutions," Apr. 2017.

[37] P. Ploton *et al.*, "Spatial validation reveals poor predictive performance of large-scale ecological mapping models," *Nat. Commun.*, vol. 11, no. 1, p. 4540, Sep. 2020, doi: 10.1038/s41467-020-18321-y.

[38] D. N. H. HORLER, M. DOCKRAY, and J. BARBER, "The red edge of plant leaf reflectance," *Int. J. Remote Sens.*, vol. 4, no. 2, pp. 273–288, Jan. 1983, doi: 10.1080/01431168308948546.

[39] Y. Wang, Z. Fang, and H. Hong, "Comparison of convolutional neural networks for landslide susceptibility mapping in Yanshan County, China," *Science of The Total Environment*, vol. 666, pp. 975–993, May 2019, doi: 10.1016/j.scitotenv.2019.02.263.

[40] B. Zhang, L. Zhao, and X. Zhang, "Three-dimensional convolutional neural network model for tree species classification using airborne hyperspectral images," *Remote Sens. Environ.*, vol. 247, Sep. 2020, doi: 10.1016/j.rse.2020.111938.

[41] X. Xu, W. Li, Q. Ran, Q. Du, L. Gao, and B. Zhang, "Multisource Remote Sensing Data Classification Based on Convolutional Neural Network," *IEEE Transactions on Geoscience and Remote Sensing*, vol. 56, no. 2, pp. 937–949, Feb. 2018, doi: 10.1109/TGRS.2017.2756851.

[42] AEP, "Legacy CCR Surface Impoundment Applicability Documentation: Glen Lyn Plant Site, Bottom Ash Pond Complex," 2024. Accessed: Sep. 29, 2026. [Online]. Available: https://docs.aep.com/docs/requiredpostings/ccr/2024/11-05-2024/GL-BAPC-ApplicabilityReport-11042024.pdf

[43] US EPA, "Final Coal Combustion Residue Impoundment Round 9 Dam Assessment Report: AEP Glen Lyn Plant, Glen Lyn, Virginia," 2011. Accessed: Sep. 30, 2026. [Online]. Available: https://www3.epa.gov/epawaste/coal/pdf/aep-glen-lyn-final.pdf

[44] EIP, "Glen Lyn Power Station." Accessed: Oct. 26, 2026. [Online]. Available: https://ashtracker.org/site/301

[45] C. F. Brown *et al.*, "Dynamic World, Near real-time global 10 m land use land cover mapping," *Sci. Data*, vol. 9, no. 1, p. 251, Jun. 2022, doi: 10.1038/s41597-022-01307-4.

[46] T. G. Farr *et al.*, "The Shuttle Radar Topography Mission," *Reviews of Geophysics*, vol. 45, no. 2, Jun. 2007, doi: 10.1029/2005RG000183.

[47] M. E. Turek *et al.*, "Global mapping of volumetric water retention at 100, 330 and 15 000 cm suction using the WoSIS database," *International Soil and Water Conservation Research*, vol. 11, no. 2, pp. 225–239, Jun. 2023, doi: 10.1016/j.iswcr.2022.08.001.

[48] Z. , H. S. , & H. G. Wan, "MODIS/Terra Land Surface Temperature/Emissivity Daily L3 Global 1km SIN Grid V061 [Data set].," 2021, doi: https://doi.org/10.5067/MODIS/MOD11A1.061.

[49] D. Yamazaki, D. Ikeshima, J. Sosa, P. D. Bates, G. H. Allen, and T. M. Pavelsky, "MERIT Hydro: A High-Resolution Global Hydrography Map Based on Latest Topography Dataset," *Water Resour. Res.*, vol. 55, no. 6, pp. 5053–5073, Jun. 2019, doi: 10.1029/2019WR024873.

[50] Y. Li, N. Wang, J. Shi, X. Hou, and J. Liu, "Adaptive Batch Normalization for practical domain adaptation," *Pattern Recognit.*, vol. 80, pp. 109–117, Aug. 2018, doi: 10.1016/j.patcog.2018.03.005.

[51] D. P. Kingma and J. Ba, "Adam: A Method for Stochastic Optimization," Jan. 2017.

[52] T. Chen and C. Guestrin, "XGBoost," in *Proceedings of the 22nd ACM SIGKDD International Conference on Knowledge Discovery and Data Mining*, New York, NY, USA: ACM, Aug. 2016, pp. 785–794. doi: 10.1145/2939672.2939785.

[53] O. Oktay *et al.*, "Attention U-Net: Learning Where to Look for the Pancreas," May 2018.

[54] I. Loshchilov and F. Hutter, "Decoupled Weight Decay Regularization," Jan. 2019.

[55] J. Cohen, "A Coefficient of Agreement for Nominal Scales," *Educ. Psychol. Meas.*, vol. 20, no. 1, pp. 37–46, Apr. 1960, doi: 10.1177/001316446002000104.

[56] H. R. Kunsch, “The Jackknife and the Bootstrap for General Stationary Observations,” *The Annals of Statistics*, vol. 17, no. 3, pp. 1217–1241, Sep. 1989, doi: 10.1214/aos/1176347265.
[57] M. Petrović and Ž. Fiket, “Environmental damage caused by coal combustion residue disposal: A critical review of risk assessment methodologies,” *Chemosphere*, vol. 299, p. 134410, Jul. 2022, doi: 10.1016/j.chemosphere.2022.134410.
[58] L. P. Padhye *et al.*, “Contaminant containment for sustainable remediation of persistent contaminants in soil and groundwater,” *J. Hazard. Mater.*, vol. 455, p. 131575, Aug. 2023, doi: 10.1016/j.jhazmat.2023.131575.
[59] S. Karim, “Multisource remote sensing data for leached metal hazard mapping at four coal combustion residual sites in Virginia, USA (2019–2025).” Accessed: Nov. 19, 2026. [Online]. Available: https://doi.org/10.5281/zenodo.23823223